\documentclass[10pt,twocolumn,letterpaper]{article}

\usepackage{cvpr}              
\usepackage{mmstyle}
\usepackage{multirow}
\usepackage[dvipsnames]{xcolor}

\usepackage[pagebackref,breaklinks,colorlinks]{hyperref}

\title{Synthesizing Physically Plausible Human Motions in 3D Scenes}

\author{
    \vspace{1mm}
    {   
        Liang Pan\textsuperscript{1}\hspace{0.5mm}
        Jingbo Wang\textsuperscript{2}\hspace{0.5mm}
        Buzhen Huang\textsuperscript{1}\hspace{0.5mm}
        Junyu Zhang\textsuperscript{1}\hspace{0.5mm}
        Haofan Wang\textsuperscript{3}\hspace{0.5mm}
        Xu Tang\textsuperscript{3}\hspace{0.5mm}
        Yangang Wang\textsuperscript{1}\footnotemark
    }
    \\
    {
        \normalsize 
        \textsuperscript{1}Southeast University\hspace{5mm}
        \quad
        \textsuperscript{2}Shanghai AI Laboratory\hspace{5mm}
        \quad
        \textsuperscript{3}Xiaohongshu Inc.\hspace{5mm}
    }
}

\begin{document}

\twocolumn[{
\renewcommand\twocolumn[1][]{#1}
\maketitle
\begin{center}
    \centering
    \captionsetup{type=figure}
    \includegraphics[width=1\textwidth]{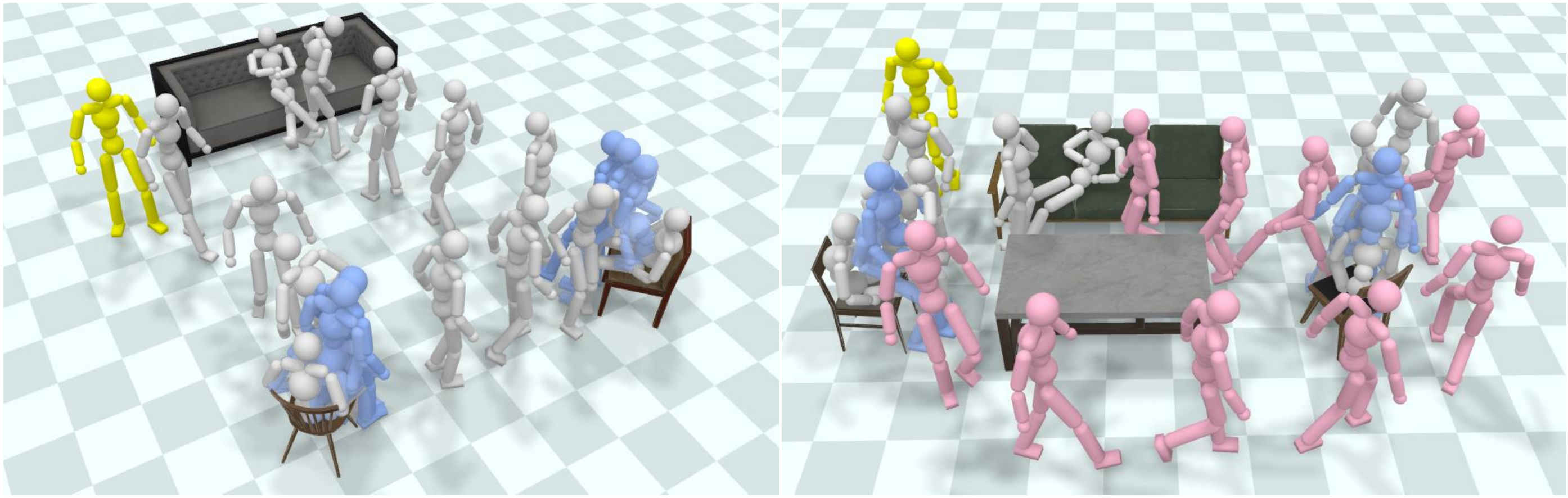}
    \captionof{figure}{We propose \textit{\textbf{InterScene}}, a novel method that generates physically plausible long-term motion sequences in 3D indoor scenes. Our approach enables physics-based characters to exhibit natural interaction-involved behaviors, such as sitting down (\textbf{\textcolor[rgb]{0.5, 0.5, 0.5}{gray}}), getting up (\textbf{\textcolor[rgb]{0.36, 0.57, 0.9}{blue}}), and walking while avoiding obstacles (\textbf{\textcolor[rgb]{0.99, 0.56, 0.67}{pink}}).}
    \label{fig:teaser}
\end{center}
}]

\renewcommand{\thefootnote}{\fnsymbol{footnote}}
\footnotetext[1]{Corresponding author. E-mail: yangangwang@seu.edu.cn. All the authors from Southeast University are affiliated with the Key Laboratory of Measurement and Control of Complex Systems of Engineering, Ministry of Education, Nanjing, China. This work was supported in part by the National Natural Science Foundation of China (No. 62076061), the Natural Science Foundation of Jiangsu Province (No. BK20220127).}

\begin{abstract}
    We present a physics-based character control framework for synthesizing human-scene interactions. Recent advances adopt physics simulation to mitigate artifacts produced by data-driven kinematic approaches. However, existing physics-based methods mainly focus on single-object environments, resulting in limited applicability in realistic 3D scenes with multi-objects. To address such challenges, we propose a framework that enables physically simulated characters to perform long-term interaction tasks in diverse, cluttered, and unseen 3D scenes. The key idea is to decouple human-scene interactions into two fundamental processes, \textbf{Inter}acting and \textbf{Nav}igating, which motivates us to construct two reusable \textbf{Con}trollers, namely \textbf{InterCon} and \textbf{NavCon}. Specifically, InterCon uses two complementary policies to enable characters to enter or leave the interacting state with a particular object (\eg, sitting on a chair or getting up). To realize navigation in cluttered environments, we introduce NavCon, where a trajectory following policy enables characters to track pre-planned collision-free paths. Benefiting from the divide and conquer strategy, we can train all policies in simple environments and directly apply them in complex multi-object scenes through coordination from a rule-based scheduler. Video and code are available at \href{https://github.com/liangpan99/InterScene}{https://github.com/liangpan99/InterScene}.
\end{abstract}

\section{Introduction}
\label{sec:intro}

Creating virtual humans with various motion patterns in daily living scenarios remains a fundamental undertaking within computer vision and graphics. While previous kinematics-based 
works~\cite{starke2019neural,wang2021synthesizing,hassan2021stochastic,wang2022towards,zhang2022couch,zhao2023synthesizing,lee2023locomotion,pi2023hierarchical,huang2023diffusion,mir23origin} have achieved long-term human motion generation in 3D indoor scenes, their models are challenging to avoid inherently physical artifacts like penetration, floating, and foot sliding. Recent physics-based works~\cite{hassan2023synthesizing,chao2021learning} began leveraging physics simulation and motion control techniques to enhance the physical realism of generated results. However, their frameworks remain constrained to simulated environments containing one isolated object since the policy trained with reinforcement learning (RL) has limited modeling capacity and thus results in gaps in synthesizing long-term sequences in complex multi-object scenes.

To bridge the gaps, our work endeavors to enable physically simulated characters to perform long-term interaction tasks in diverse, cluttered, and unseen 3D scenes. The key insight of our framework is to construct two reusable controllers, \ie, \textbf{InterCon} and \textbf{NavCon}, for learning comprehensive motor skills referring to the fundamental processes in human-scene interactions. InterCon learns skills that require environmental affordances, such as sitting on a chair and rising from seated positions. NavCon considers environmental constraints to control characters' locomotion following collision-free paths. The strengths lie in 1) our framework decouples long-term interaction tasks into a scheduling problem of two controllers; 2) both controllers can be trained in relatively simple environments without relying on costly 3D scene data; 3) trained controllers can directly generalize to complex multi-object scenes without additional training. 

Although existing works~\cite{chao2021learning,hassan2023synthesizing} presented controllers for executing interaction tasks, they cannot be applied in multi-object scenarios due to incomplete interaction modeling. To address this issue, our InterCon employs two complementary control policies to learn complete interaction skills. Different from previous works~\cite{chao2021learning,hassan2023synthesizing}, InterCon not only involves reaching and interacting with the object but also includes leaving the interacting object. As illustrated in Fig~\ref{fig:teaser}~(left), leveraging the two policies guarantees InterCon is a closed-loop controller that realizes interactions with multiple objects. In cluttered environments with obstacles like Fig~\ref{fig:teaser}~(right), we introduce NavCon to enable characters to navigate and avoid obstacles. InterCon and NavCon provide complete interaction skills, including sitting, getting up, and obstacle-free trajectory following. Thus, we can leverage a finite state machine to schedule the two controllers, allowing characters to perform long-term interaction tasks in complex 3D scenes without additional training.

We train all policies by goal-conditioned reinforcement learning and adversarial motion priors (AMP)~\cite{peng2021amp}. It is challenging to train InterCon stably because the policy needs to coordinate the character's fine-grained movement in relation to the object, and the reward is sparse. Previous work~\cite{hassan2023synthesizing} conditions the discriminator on the scene context to construct a dense reward. In this work, we propose interaction early termination to achieve stable training via balancing data distribution in training samples. To ensure that our InterCon can generalize to unseen objects, various objects with diverse shapes are required for the training~\cite{hassan2023synthesizing}. However, the get-up policy relies on various seated poses in plausible contact with objects (\eg, without floating and penetration), which is difficult to obtain. To tackle this issue, we introduce seated pose sampling, where a trained sit policy will generate plausible sitting poses without incurring additional costs of motion capture.

In summary, our main contributions are:
\begin{enumerate}
    \vspace{1mm}
    \item We propose a system that enables physically simulated characters to perform long-term interaction tasks in diverse, cluttered, and unseen 3D scenes.
    \vspace{1mm}
    \item We propose two reusable controllers for modeling interaction and navigation to decouple challenging human-scene interactions.
    \vspace{1mm}
    \item We leverage a rule-based scheduler that enables users to create human-scene interactions through intuitive instructions without additional training.
\end{enumerate}

\section{Related Work}
\label{sec:relatedwork}

\begin{figure*}
  \centering
   \includegraphics[width=1.0\linewidth]{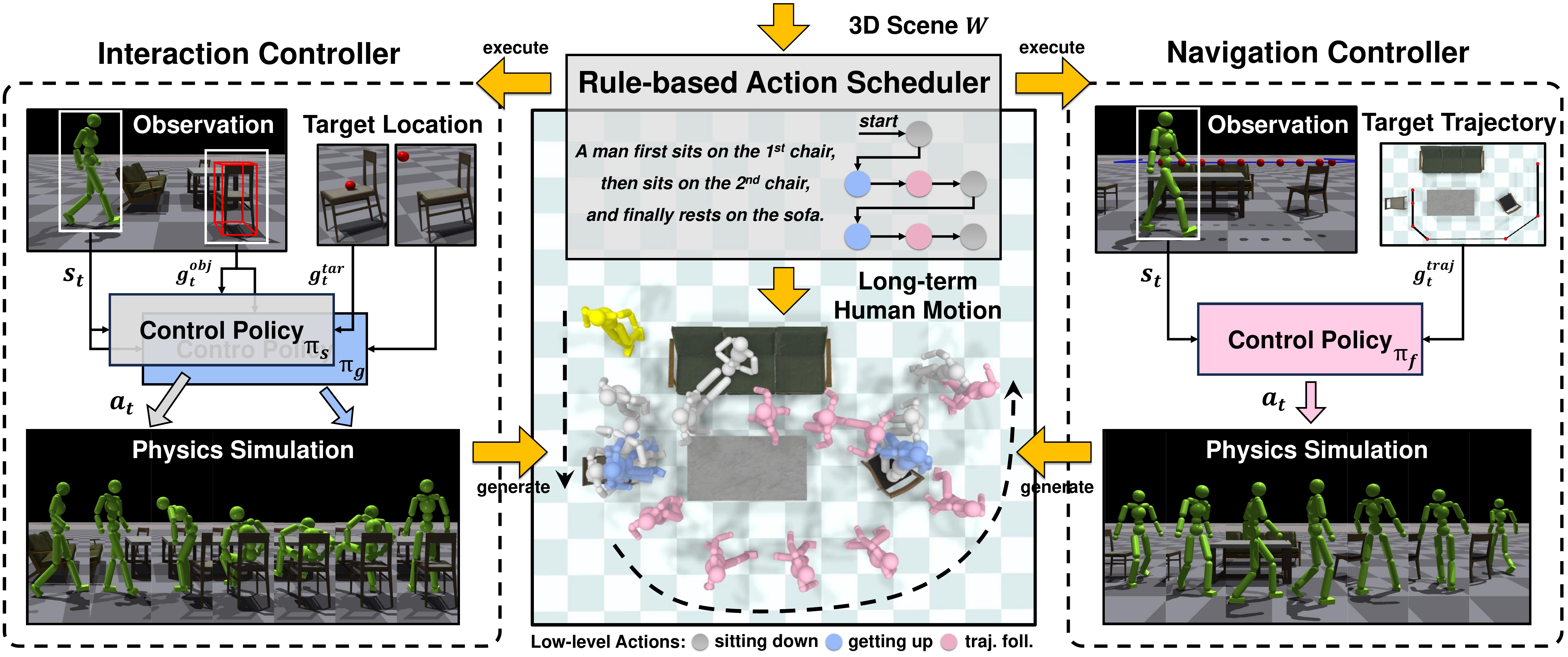}

   \caption{\textbf{System overview.} Given a multi-object 3D scene, our goal is to synthesize long-term motion sequences by controlling a physics-based character to perform a series of scene interaction tasks. First, our system employs an interaction controller to provide two primary actions, \ie, sitting down and getting up. Second, we introduce a navigation controller to acquire another action, \ie, collision-free trajectory following. Finally, a rule-based action scheduler is exploited to obtain outputs by organizing reusable low-level actions according to user-designed instructions.}
   \label{fig:pipeline}
\end{figure*}

\noindent \textbf{Human-Scene Interaction} Creating virtual characters capable of interacting with surrounding environments is widely explored in computer vision and graphics. One of the streams of methods is to build data-driven kinematic models leveraging large-scale motion capture datasets~\cite{mahmood2019amass}. Phase-based neural networks~\cite{holden2017phase,starke2019neural,starke2020local,starke2022deepphase} are widely used in generating natural and realistic human motions. Holden \etal~\cite{holden2017phase} propose PFNNs to produce motions where characters adapt to uneven terrain. Strake \etal~\cite{starke2019neural} extend the idea of phase variables to generate motions in human-scene interaction scenarios such as sitting on chairs and carrying boxes. Strake \etal~\cite{starke2020local} propose a local motion phase based model for synthesizing contact-rich interactions. An alternative approach uses generative models like conditional variance autoencoder (cVAE) and motion diffusion model (MDM)~\cite{tevet2022human,zhou2023emdm,xu2023interdiff,li2023object} to model human-scene interaction behaviors. Wang \etal~\cite{wang2021synthesizing} present a hierarchical generative framework to synthesize long-term 3D motion conditioning on the 3D scene structure. Hassan \etal~\cite{hassan2021stochastic} present a stochastic scene-aware motion generation framework using two cVAE models for learning target goal position and human motion manifolds. Wang \etal~\cite{wang2022towards} present a framework to synthesize diverse scene-aware human motions. Taheri \etal~\cite{taheri2022goal} and Wu \etal~\cite{wu2022saga} design similar frameworks to generate whole-body interaction motions. Most recent works adopt reinforcement learning (RL) to develop control policies for motion generation. MotionVAE~\cite{ling2020character} is a most representative work that proposes a new paradigm for motion generation based on RL and generative models. Zhang \etal~\cite{zhang2022wanderings} extend MotionVAE to synthesize diverse digital humans in 3D scenes. Zhao \etal~\cite{zhao2023synthesizing} present interaction and locomotion policies to synthesize human motions in 3D indoor scenes. Lee \etal~\cite{lee2023locomotion} also use reinforcement learning and motion matching to solve the locomotion, action, and manipulation task in 3D scenes. In this work, we aim to synthesize 3D motions of interacting with everyday indoor objects (\eg, chairs, sofas, and stools). Most relevant previous works are phase-based neural networks~\cite{starke2019neural}, cVAE-based generative models~\cite{hassan2021stochastic,wang2021synthesizing,wang2022towards}, and RL-based methods~\cite{zhao2023synthesizing,lee2023locomotion}.

\noindent \textbf{Physics-based Character Control} Physics-based methods focus on developing motion control techniques to animate characters in physics simulators~\cite{makoviychuk2021isaac,todorov2012mujoco}. In recent years, various motion imitation (or motion tracking) based methods have been established to enable simulated characters to imitate diverse, challenging, and natural motor skills. Liu~\etal~\cite{liu2010sampling,liu2015improving} propose sampling-based methods equipped with Covariance Matrix Adaptation (CMA)~\cite{hansen2006cma}. DeepMimic~\cite{peng2018deepmimic} adopts deep reinforcement learning (DRL) to train policy neural networks. Model-based methods~\cite{fussell2021supertrack,yao2022controlvae,won2022physics} use supervised learning to train policies efficiently. Tracking-based methods have trained control policies to animate simulated characters for carrying box~\cite{xie2023hierarchical,chao2021learning}, sports~\cite{10.1145/3197517.3201315,10.1145/3450626.3459817}, learning skills from videos~\cite{10.1145/3592408,neuralmocon,10.1145/3272127.3275014}. However, motion tracking needs reference motions to implement an imitation objective. It can be challenging to obtain desired reference motions when policies are applied to perform new tasks that require diverse skills. Recently, Peng~\etal~\cite{peng2021amp} introduce Generative Adversarial Imitation Learning (GAIL)~\cite{10.5555/3157382.3157608} to the character animation field and present Adversarial Motion Priors (AMP)~\cite{peng2021amp}. AMP replaces the complex tracking-based objectives with a motion discriminator trained on large unstructured datasets. AMP-based works have achieved impressive results in both motion imitation and motion generation. A series of papers~\cite{peng2022ase,juravsky2022padl,tessler2023calm,dou2023c} use AMP to learn latent skill embeddings from large motion datasets and then train a high-level policy to reuse the embeddings to solve downstream tasks. Luo~\etal~\cite{luo2023perpetual} propose a motion imitator that can track a large-scale motion sequences. Rempe~\etal~\cite{rempe2023trace} build a pedestrian animation system using controllers trained with AMP to generate pedestrian locomotion. InterPhys~\cite{hassan2023synthesizing} first extends the AMP framework to human-scene interaction tasks, such as sitting on chairs, lying on sofas, and carrying boxes. UniHSI~\cite{xiao2023unified} involves Large Language Models to drive simulated characters to perform scene interactions. In this work, we focus on synthesizing long-term interactions in cluttered 3D scenes.

\section{Method}
\label{sec:method}

\subsection{Preliminaries}
\label{sec:preliminaries}

\noindent \textbf{Motion Synthesis Paradigm} We achieve motion synthesis through physics-based character control. To train policies that enable characters to perform tasks in a life-like manner, we leverage the goal-conditioned RL framework of Adversarial Motion Priors (AMP)~\cite{peng2021amp}. At each time step $t$, the policy $\pi(a_t|s_t,g_t)$ predicts an action $a_t$ based on the character state $s_t$ and the task-specific goal state $g_t$. Applied that action, the environment transitions to the next state $s_{t+1}$ based on its dynamics $p(s_{t+1}|s_t,a_t)$. Then it receives a scalar reward $r_t$ computed by $r = w^Gr^G + w^Sr^S$, where $r^G$ is a task reward designed using human knowledge, and $r^S$ is a naturalness reward modeled by a motion discriminator. The policy is trained to maximize the expected discounted return $J(\pi) = \mathbb{E}_{p(\tau | \pi)}\left[ \sum_{t=0}^{T-1}\gamma^{t} r_{t} \right]$, where $T$ is the horizontal length and $\gamma \in [0, 1]$ defines the discount factor.

\vspace{2mm}
\noindent \textbf{Character Model and State Representation} The character has $15$ rigid bodies, $12$ movable internal joints, and $28$ DoF actuators. We exploit proportional derivative (PD) controllers to convert the action $a \in \mathbb{R}^{28}$ into torques to actuate the internal joints. The root joint is not controllable. The character state $s \in \mathbb{R}^{223}$ input to the policy network is in the maximal coordinate system, including:
\begin{itemize}
    \item Root height $s^{rh} \in \mathbb{R}^1$
    \item Root rotation $s^{rr} \in \mathbb{R}^6$
    \item Root linear velocity $s^{rv} \in \mathbb{R}^3$ 
    \item Root angular velocity $s^{ra} \in \mathbb{R}^3$ 
    \item Positions of other bodies $s^{jp} \in \mathbb{R}^{14 \times 3}$ 
    \item Rotations of other bodies $s^{jr} \in \mathbb{R}^{14 \times 6}$ 
    \item Linear velocities of other bodies $s^{jv} \in \mathbb{R}^{14 \times 3}$ 
    \item Angular velocities of other bodies $s^{ja} \in \mathbb{R}^{14 \times 3}$ 
\end{itemize}
The height and rotation of the root are recorded in the world coordinate frame, and other terms are recorded in the character's local coordinate frame. We use $6$D rotation representations~\cite{zhou2019continuity}. The state input to the discriminator $s \in \mathbb{R}^{125}$ is in the reduced coordinate system. Please refer to our publicly released code for more details.

\vspace{2mm}
\noindent \textbf{Location Task} \hspace{0.5mm} As illustrated in Fig~\ref{fig:pipeline}, our system contains three control policies: sit policy $\pi_s$, get-up policy $\pi_g$, and trajectory following policy $\pi_f$. Although they are trained for different scene interaction tasks, \ie., sitting on chairs, getting up from seated states, and following trajectories, we can implement the policy training in a generic \textit{location task}. The task objective is for the character to move its root to a target location. For instance, the target location can be formulated as a point on a chair seat that we expect the character to sit on or a point in a trajectory that the character should approach. Then, we carefully design the task settings for training each policy. In the subsequent Sec~\ref{sec:method_intercon} and Sec~\ref{sec:method_navicon}, we will describe them in detail. Besides, we can detect whether a task is completed by measuring the distance between the root and target location, which the rule-based action scheduler uses to perform action transition.

\begin{figure}[t]
  \centering
   \includegraphics[width=1.0\linewidth]{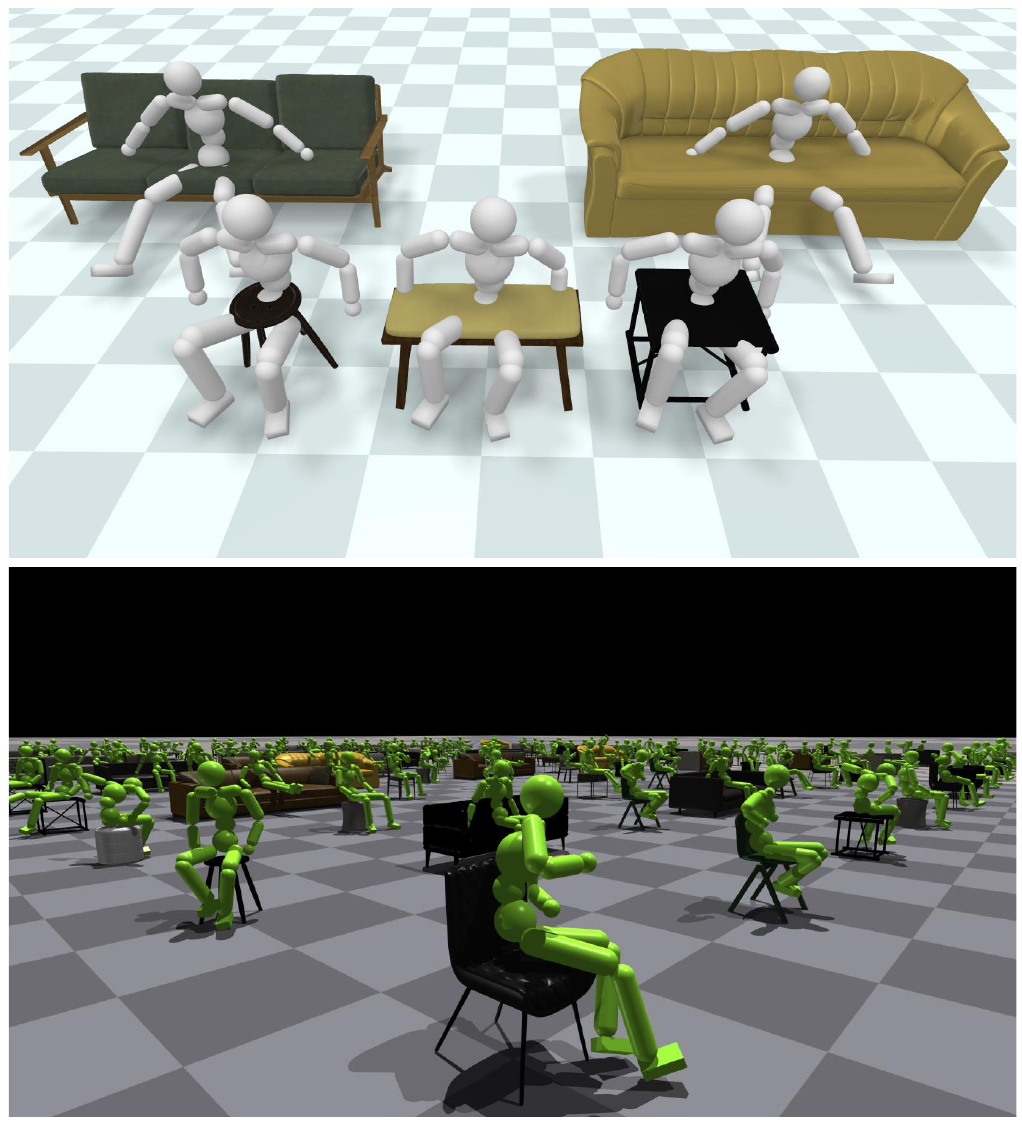}
   \caption{Seated poses sampled from the reference dataset (upper row) and generated by pre-trained sit policy (lower row).}
   \label{fig:penetration}
\end{figure}

\subsection{System Overview}
\label{sec:method_overview}

As illustrated in Fig~\ref{fig:pipeline}, our complete system integrates two reusable controllers, \ie, \textbf{Inter}action \textbf{Con}troller (InterCon) and \textbf{Nav}igation \textbf{Con}troller (NavCon), serving as low-level executors, and a Rule-based Action Scheduler, serving as a high-level planner to schedule two executors to synthesize human motions according to user instructions. InterCon consists of two control policies: sit policy $\pi_{s}$ and get-up policy $\pi_{g}$. Each policy is conditioned on the target object features $g_t^{obj}$ and the target location $g_t^{tar}$ of the character's root. NavCon contains a trajectory following policy $\pi_{f}$ conditioned on the target trajectory features $g_t^{traj}$. These input conditions can be seen as explicit control signals.

We use the cluttered 3D scene $W$ shown in Fig~\ref{fig:pipeline}, which contains three interactable objects $W=\{w_1,w_2,w_3\}$, as an example to describe the workflow of using our system to generate long-term interactions. Based on three control policies, we provide users with three reusable actions: sitting down $k_s$, getting up $k_g$, and trajectory following $k_f$. To synthesize human motions described by ``A man first sits on the $1^{\text{st}}$ chair, then sits on the $2^{\text{nd}}$ chair, and finally rests on the sofa'', the user needs to construct the following instruction:
\begin{equation}
  \begin{aligned}
      I = \{ &(k_s,w_1), \\
      &(k_g,w_1), (k_f,h_1), (k_s,w_2), \\
      &(k_g,w_2), (k_f,h_2),(k_s,w_3) \},
  \end{aligned}
  \label{eq:instruction}
\end{equation}
where $(k_s/k_g,w)$ denotes that the character performs the action $k_s$ to sit on the object $w$ or performs the action $k_g$ to get up from the object $w$, and $(k_f,h)$ denotes that the character walks along an obstacle-free trajectory $h$ that can be either generated by the A* path planning algorithm~\cite{hart1968formal} or defined by the user. The rule-based action scheduler translates this instruction into a sequence of explicit control signals and then schedules low-level policies to execute the instruction. It is also responsible for performing action transitions. For instance, an action will be terminated when the overlapping time between the character's root and its current target outperforms a fixed time.

\subsection{Interaction Controller}
\label{sec:method_intercon}

Given a target object $w_i \in W$, our InterCon aims to enable characters to enter or leave the interacting state with the object. Previous methods~\cite{chao2021learning,hassan2023synthesizing} mainly focus on developing the former ability for the character, overlooking the latter's importance for long-term interaction tasks. By incorporating a new skill of getting up, the controller allows the character to transition from a seated state to 
a standing state, which provides an opportunity to perform the next new interaction task.

\vspace{2mm}
\noindent \textbf{Task-specific Goal State} As illustrated in Fig~\ref{fig:pipeline}, we construct InterCon using two separate policies. They share the same character state $s_t$, task-specific goal state $g_t=\{g_t^{obj},g_t^{tar}\}$, and network structure. We input the target object's features $g_t^{obj} \in \mathbb{R}^{3+6+2+24}$ that contain the object's position and rotation, the horizontal vector of its facing direction, and $8$ vertices of its bounding box to the policy to enable it to be aware of the target object state. We also follow~\cite{hassan2023synthesizing} to condition the discriminator on the object state $g_t^{obj}$, which is crucial for policy to effectively learn how to coordinate the movement of a character concerning the target. In addition, we add explicit target location $g_t^{tar} \in \mathbb{R}^{3}$ into the goal features $g_t$. All these goal features are recorded in the character's local frame. The target location of the sit task is generated before training and placed $10$ cm above the center of the top surface of the object seat. The target location of the get-up task is computing online according to the states of foot bodies. Please refer to our publicly released code for more details.

\vspace{2mm}
\noindent \textbf{Motion and Object Datasets} We use $111$ motion sequences from the SAMP dataset~\cite{hassan2021stochastic} containing multiple behaviors (walking, sitting, and getting up) and diverse human-object configurations. We manually split the motion dataset into two subsets used to train the sit and get-up policies, respectively. To construct the object dataset, we select $40$ straight chairs, $40$ low stools, and $40$ sofas from the 3D-Front object dataset~\cite{fu20213d} and randomly divide $30$ as the training set and $10$ as the testing set. We coarsely fit objects to human motions since we condition the discriminator on the object state.

\begin{figure}[t]
  \centering
   \includegraphics[width=1.0\linewidth]{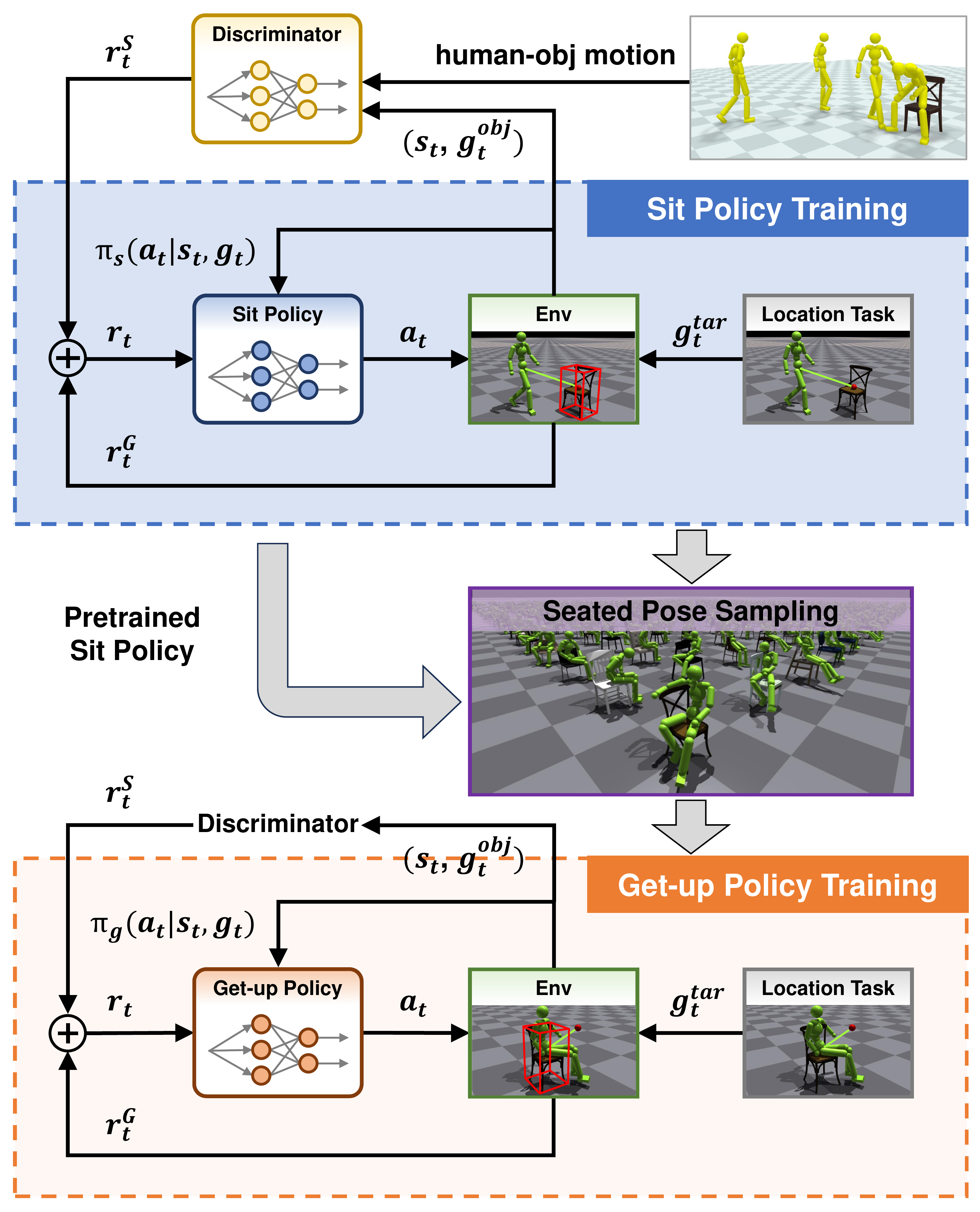}

   \caption{\textbf{The process for training the sit and get-up policies} consists of three steps. \textbf{1) Sit Policy Training:} Inspired by~\cite{hassan2023synthesizing}, we first extend the standard AMP framework with several improvements to train a robust sit policy. \textbf{2) Seated Pose Sampling:} We tackle the issue of lacking high-quality seated human poses adapted to various object shapes by using the pre-trained sit policy to generate numerous seated poses randomly. \textbf{3) Get-up Policy Training:} We adopt a similar method to train the get-up policy. At the beginning of each training episode, the character will be initialized to a seated state sampled from the previously synthesized database.}
   \label{fig:intercon_training}
\end{figure}

\vspace{2mm}
\noindent \textbf{Initialization} At the beginning of each episode, we need to initialize them appropriately to facilitate the training process. For the sit task, we utilize Reference State Init (RSI)~\cite{peng2018deepmimic}  to sample interaction states from the previously fitted dataset randomly. As shown in Fig~\ref{fig:penetration}, partial frames in a seated state will penetrate with objects, which will hurt physics simulation. Therefore, we pass these frames during RSI. We also follow~\cite{hassan2023synthesizing} to encourage the character to execute the sit task from a wide range of initial configurations by randomizing the position and rotation of objects. For the get-up task, we should start training from seated states. However, it is difficult to obtain reasonable states from reference. Moreover, constructing an accurately aligned dataset is also costly. Thus, we build a synthetic database, illustrated in Fig~\ref{fig:penetration}, consisting of plenty of seated poses with high-quality contact. It will be further discussed in subsequent paragraphs.

\vspace{2mm}
\noindent \textbf{Reset and Early Termination} An episode terminates after a fixed episode length or when early termination conditions have been triggered. The episode length is set to 10 seconds. We use fall detection as a basic condition. In order to improve sampling efficiency, we introduce interaction early termination (IET). IET triggers when the accumulative overlapping time between the character's root $x^{root}_t$ and the target location $g_t^{tar}$ exceeds a fixed threshold. Such a simple mechanism can effectively assist the RL training process.

\vspace{2mm}
\noindent \textbf{Training Scheme of Two Policies} The training process is detailed in Fig~\ref{fig:intercon_training}. There is a dependency graph among the sitting and getting up behaviors. We need to initialize the character to a plausible seated state at the beginning of each episode to train the get-up policy. However, collecting such a diverse set of initial states is hard and costly. Thus, our training scheme starts  upstream in the dependency graph to train a sit policy first. Subsequently, we directly employ it to perform seated pose sampling to obtain high-quality interaction states with no floating and penetration. We collect $300$ poses for each training object. In total, we generate $300 \times 90$ poses to form a synthetic database. A snapshot of partial data is shown in Fig~\ref{fig:penetration}. Then, we train the get-up policy from scratch and keep most of the settings. Such a database is the key to successfully training the get-up policy.

\vspace{2mm}
\noindent \textbf{Task Rewards} Given a target location $g_t^{tar}$ and a target velocity $g_t^{vel}=1.5 \text{m/s}$, the task reward for sit is defined as:

\begin{equation}
    r^G_t=\left\{
    \begin{aligned}
    0.7 \ r_t^{near} &+ 0.3 \ r_t^{far}, \left \| x_t^* - x^{root}_t \right \|^2 > 0.5 \\
    0.7 \ r_t^{near} &+ 0.3, \text{otherwise}
    \end{aligned}
    \right.
\label{eq:sit_total_reward}
\end{equation}
\begin{equation}
    \begin{aligned}
    r_t^{far} &= \text{exp} \big ( -2.0 \left \| g_t^{vel} - d_t^* \cdot \dot{x}^{root}_t \right \|^2 \big )
    \end{aligned}
\label{eq:far_reward}
\end{equation}
\begin{equation}
r_t^{near} = \text{exp}\big(-10.0 \left \| g_t^{tar} - x_t^{root} \right \|^2 \big)
\label{eq:near_reward}
\end{equation}
where $x^{root}_t$ is the position of the character's root,  $\dot{x}^{root}_t$ is the linear velocity of the character's root, $x^*_t$ is the position of the object, $d^*_t$ is a horizontal unit vector pointing from $x^{root}_t$ to $x^*_t$, $a \cdot b$ represents vector dot product. We only use the $r_t^{near}$ as the task reward for the get-up task.

\subsection{Navigation Controller}
\label{sec:method_navicon}

\noindent

While InterCon has been able to generate long-term interactions, it cannot avoid obstacles. To learn a necessary skill in cluttered environments, \ie, collision-free navigation, we introduce NavCon to our system, which contains a trajectory following policy $\pi_{f}(a_t|s_t,g_t^{traj})$~\cite{rempe2023trace} and off-the-shelf path planning algorithms, as shown in Fig~\ref{fig:pipeline}. Incorporating a module for obtaining collision-free locomotion has been widely used in previous works~\cite{wang2022towards,hassan2021stochastic,zhao2023synthesizing}. However, kinematics-based methods produce artifacts like foot skating and penetration. Physics-based Pacer~\cite{rempe2023trace} can generate physically plausible gaits and contacts. In this work, we apply this technique in the character-scene interaction field to solve the navigation problem.

\vspace{2mm}
\noindent \textbf{Task-specific Goal State} We construct the goal features using a short future path $g_t^{traj} \in \mathbb{R}^{10 \times 2}$ consisting of a sequence of $2$D target positions of the character's root for the next $1.0$ seconds sampled at $0.1$s intervals.

\vspace{2mm}
\noindent \textbf{Motion and Trajectory Datasets} We use $\sim 200$ sequences from the AMASS dataset~\cite{mahmood2019amass}. Target trajectories used for training are procedurally generated. A complete trajectory $\tau=\{ p_0^{\tau}, ..., p_{T-1}^{\tau}, p_{T}^{\tau}   \}$ is modeled as a set of $2$D points with a fixed 0.1 seconds time interval. At each simulation time step $t$, we query 10 points $\{ p_{t}^{\tau}, ..., p_{t+9}^{\tau} \}$ in the future 1.0 seconds from the complete trajectory $\tau$ by interpolating.

\vspace{2mm}
\noindent \textbf{Other Settings} We terminate the training episode when the character falls or deviates too far from the established trajectory. Characters are initialized using RSI. Trajectories will be re-generated when environments reset. The task reward $r_t^G$ measures how far away the root $x^{root}_t$ is on the horizontal plane from the desired location $p^{\tau}_t \in \tau$:
\begin{equation}
    r^G_t = \text{exp} \big ( - 2.0 \left \| x^{root}_t - p^{\tau}_t \right \|^2  \big ).
    \label{eq:traj_reward}
\end{equation}

\section{Experiment}
\label{sec:experiment}

\begin{table}
  \centering
  \resizebox{1.0\linewidth}{!}{
    \begin{tabular}{ | c | ccc | }
      \noalign{\hrule height 1pt}
      \textbf{Method} & \textbf{Success Rate (\%)} & \textbf{Execution Time (s)}  & \textbf{Error (mm)} \\
      \hline
      Chao \etal~\cite{chao2021learning} & 17.0 & -- & -- \\
      AMP~\cite{peng2021amp} & 87.5 & 4.0 & \textbf{36.5} \\
      InterPhys~\cite{hassan2023synthesizing} & 93.7 & 3.7 & 90.0  \\
      Ours & \textbf{98.8} & \textbf{2.5} & 36.8 \\
      \noalign{\hrule height 1pt}
    \end{tabular}
  }
  \caption{Comparisons with physics-based methods on sit task.}
  \label{tab:sit_task}
\end{table}

\begin{table}
  \centering
  \resizebox{1.0\linewidth}{!}{
    \begin{tabular}{| c | ccc | }
      \noalign{\hrule height 1pt}
      \textbf{IET Step} & \textbf{Success Rate (\%)} & \textbf{Execution Time (s)}  & \textbf{Error (mm)} \\
      \hline
       30 & \textbf{98.8} & \textbf{2.5} & 36.8 \\
       60 & 93.9 & 3.1 & \textbf{34.9} \\
       90 & 92.3 & 3.3 & 36.3 \\
       $\times$ & 87.5 & 4.0 & 36.5 \\
      \noalign{\hrule height 1pt}
    \end{tabular}
  }
  \caption{Metrics of sit policies trained with various early termination settings. $\times$ means that the model is trained without IET.}
  \label{tab:IET_ablation}
\end{table}

\begin{figure}[t]
  \centering
   \includegraphics[width=1.0\linewidth]{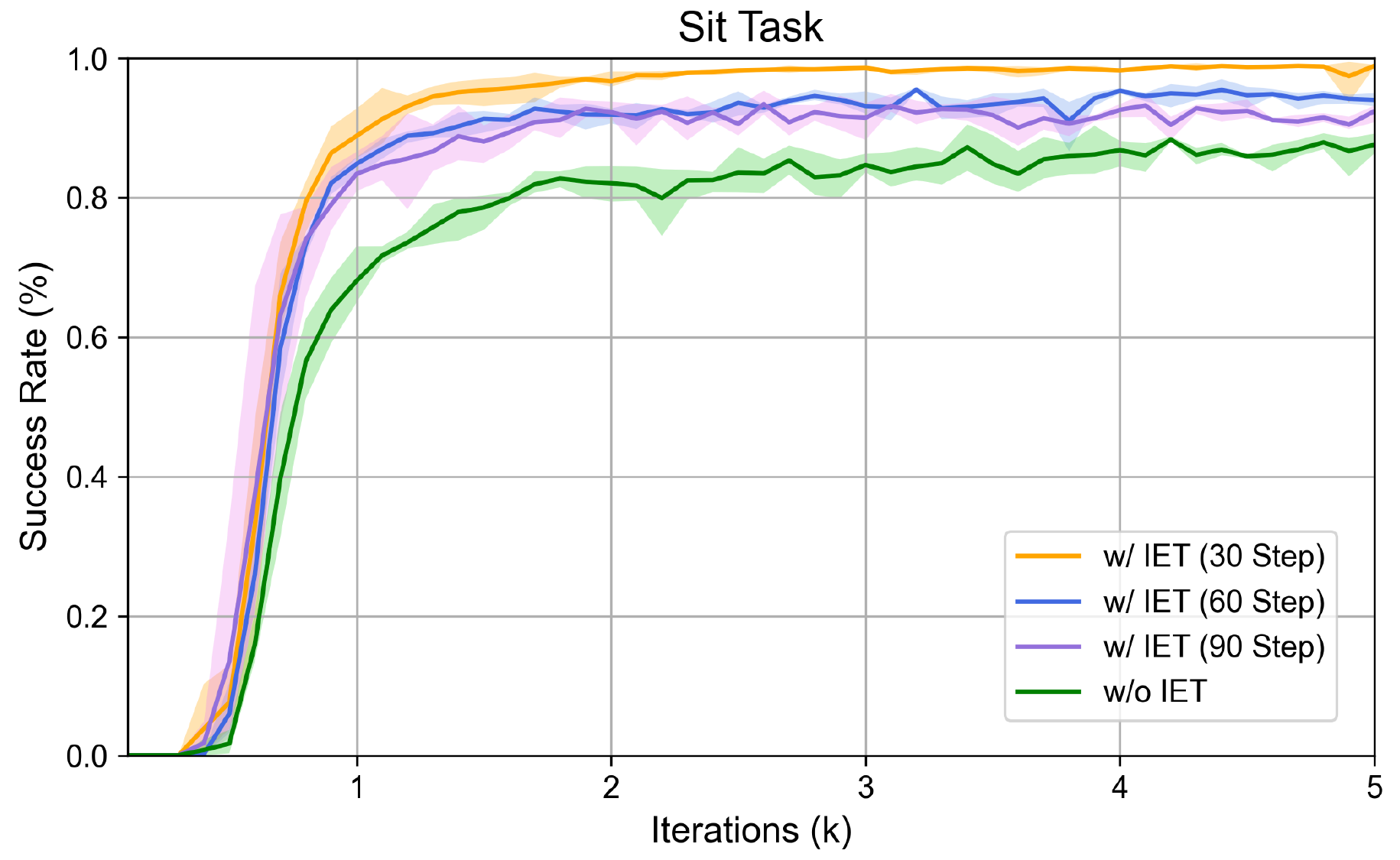}
   \caption{Performance curves of sit policies trained with various early termination settings. Colored regions denote the fluctuation range over 3 models.}
   \label{fig:performance_curve}
\end{figure}

\begin{figure*}
  \centering
   \includegraphics[width=1.0\linewidth]{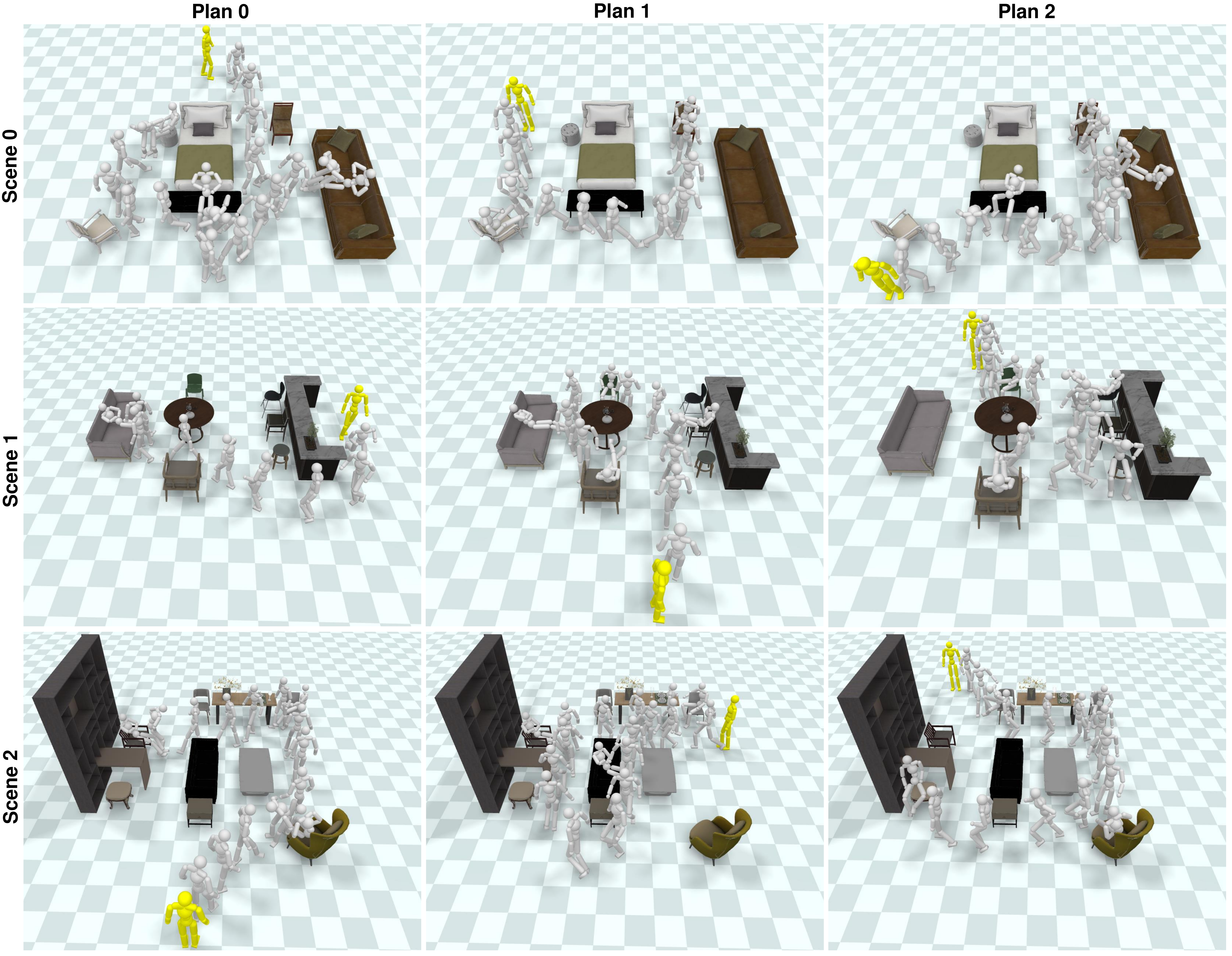}
   \caption{Our complete system successfully generates long-term motion sequences in three challenging 3D indoor scenes.}
   \label{fig:scene}
\end{figure*}

\subsection{Individual Tasks}
\label{subsec:eval_individual_tasks}
The interaction controller has two policies responsible for performing sit and get-up tasks. We first quantitatively demonstrate the effectiveness of each policy in its corresponding task. Then, we conduct ablation studies on the interaction early termination (IET). All experimental results are collected in single-object environments in which the object is randomly sampled from the object testing set. We follow~\cite{hassan2023synthesizing} to comprehensively evaluate the task performance by measuring success rate, execution time, and error. A trial will be determined to be successful if the Euclidean distance between the character's root and its target location is less than $20$ cm. To mitigate randomness, we train $3$ models initialized with different random seeds and evaluate each model using $4096$ trials for each experiment setting. Thus, all metrics are averaged over $3 \times 4096$ trials.

\vspace{2mm}
\noindent \textbf{Sit Task} Table~\ref{tab:sit_task} summarizes the quantitative comparisons between our sit policy and existing physics-based methods. We use AMP~\cite{peng2021amp} as baseline to train a policy where IET is not employed, and the discriminator's obs does not include object states. Compared with InterPhys~\cite{hassan2023synthesizing}, our training involves more detailed object information and a new termination strategy (the IET step parameter is set to $30$). During testing, the object is placed $[1, 5]$ m away from the character with a random orientation. Experimental results show that our method achieves a high success rate of $98.8\%$ and significantly outperforms others. Using IET can focus the distribution of samples collected during RL training on approaching the object and sitting down. We demonstrate that IET can improve performance effectively.

\vspace{2mm}
\noindent \textbf{Get-up Task} We first use a pre-trained sit policy to collect $300$ seated states of a given testing object. These states contain various facing directions and body poses. When testing, the character is initialized to a random seated state. Our get-up policy can achieve a success rate of $94.6\%$, which demonstrates its effectiveness.

\vspace{2mm}
\noindent \textbf{Ablation Studies on IET} For a more in-depth analysis of the impact of IET, we construct $4$ variants equipped with various early termination settings, including \textit{w/o IET} and \textit{w/ IET} with varied step parameters. Metrics in Table~\ref{tab:IET_ablation} suggest that task performance gradually improves as the step parameter decreases. Fig~\ref{fig:performance_curve} also illustrates that using a small step parameter can effectively reduce the fluctuation margin to stabilize the training process.

\begin{table}
  \centering
  \resizebox{0.6\linewidth}{!}{
    \begin{tabular}{| c | ccc | }
      \noalign{\hrule height 1pt}
      \textbf{3D Scene} & \textbf{Plan 0} & \textbf{Plan 1}  & \textbf{Plan 2} \\
      \hline
       0 & 92.1 & 99.2 & 82.0 \\
       1 & 98.4 & 88.2 & 90.6 \\
       2 & 57.0 & 71.8 & 57.8 \\
      \noalign{\hrule height 1pt}
    \end{tabular}
  }
  \caption{Success rate metrics of the complete system in final application environments. Each synthesis plan is tested by $128$ trials.}
  \label{tab:scene_demo}
\end{table}

\subsection{Long-term Motion Synthesis}
\label{subsec:eval_scene_demo}

The ultimate goal of this work is to synthesize long-term motions involving multiple actions in diverse and cluttered 3D scenes. We quantitatively and qualitatively evaluate the effectiveness of our complete system in multi-object environments. $3$ synthetic scenes are constructed using unseen objects from the 3D-Front dataset~\cite{fu20213d}. Each scene is populated with $5 \sim 6$ interactable objects and some obstacles. For each scene, we design $3$ plans manually, each of which contains a sequence of actions for the character to perform.

\vspace{2mm}
\noindent \textbf{Experimental Results} As shown in Fig~\ref{fig:scene}, our complete system can successfully synthesize desired long-term motions. It validates the proposed divide and conquer idea that the policies trained in simple environments can generalize to unseen complex 3D scenes. We also provide quantitative metrics of success rate for all plans in Table~\ref{tab:scene_demo}, which shows that performance is unstable and highly relevant to the complexity of the scene and plan. 

\begin{figure}[t]
  \centering
   \includegraphics[width=1.0\linewidth]{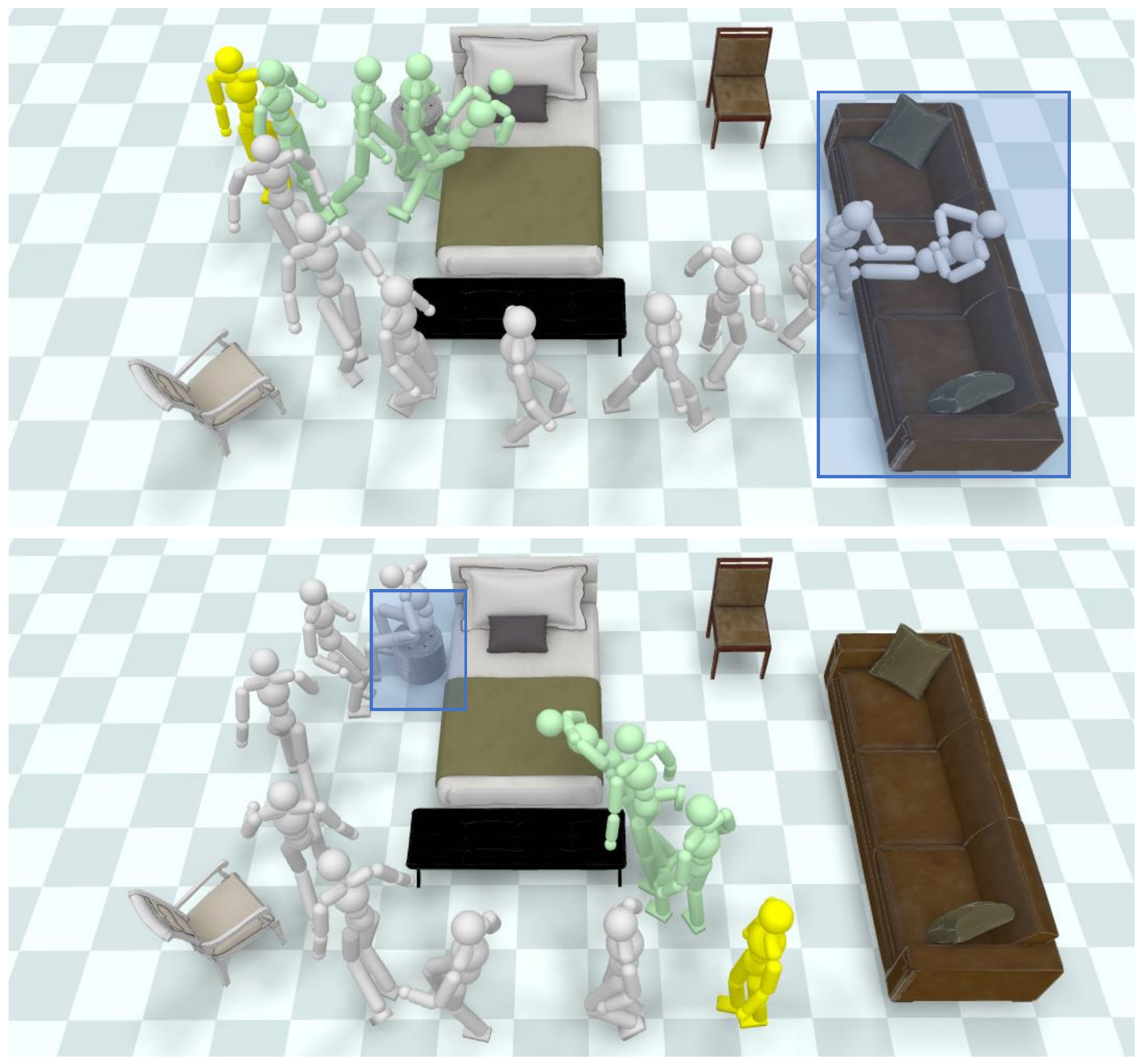}
   \caption{Comparisons of the complete system (gray) and a variant without NavCon (green). Blue rectangles denote target objects.}
   \label{fig:scene_ablation}
\end{figure}

\vspace{2mm}
\noindent \textbf{Ablation Studies on NavCon} Our system relies on NavCon to avoid obstacles. We validate its importance by qualitatively comparing motions generated by the complete system and a variant without NavCon. As illustrated in Fig~\ref{fig:scene_ablation}, given the same target, the former can successfully control the character to interact with the object. The latter gets the character stuck by obstacles because InterCon tends to follow the shortest path to the target. 

\section{Limitations and Future Work}
\label{sec:limiations}

This work aims to design a physics-based animation system to synthesize motions in complex 3D scenes and lets the system come true. However, many valuable problems remain that should be investigated in future work. Firstly, since the system is built on AMP, policies are prone to master a small subset of behaviors depicted in the motion dataset. To tackle this issue, we can explore how to introduce conditional generation capabilities~\cite{peng2022ase,dou2023c} into the system to improve skill diversity. Secondly, the system cannot handle with abnormal situations that would occur when executing tasks in complex scenes. For example, NavCon tracks a pre-planned and fixed path. Due to tracking errors, it is inevitable for the character to deviate from the path and collide with obstacles. This is also the main reason why the performance is unstable in Table~\ref{tab:scene_demo}. Creating a closed-loop system with periodic re-planning features will be necessary to improve the robustness. Thirdly, our policy only observes the state of the isolated target object and is unaware of the surroundings. Future work should explore incorporating more global representations of the cluttered environment. In addition, the real world has more complex environments, such as dynamic scenes, uneven terrains, multi-floor buildings, and narrow spaces, which are not considered in this work. It would be exciting to create autonomous agents living in a physically simulated environment with the above characteristics.

\section{Conclusion}
\label{sec:conclusion}

In this paper, we presented a physics-based animation system to synthesize long-term human motions in diverse, cluttered, and unseen 3D scenes. This is achieved by jointly using two reusable controllers, \ie, InterCon and NavCon, as well as a rule-based action scheduler, to decompose the human-scene interactions. We introduced some training techniques to train the policies of InterCon successfully. We qualitatively and quantitatively evaluated our complete system's long-term motion generation ability. Our physically simulated characters realistically and naturally presented interaction and locomotion behaviors.

{
    \small
    \bibliographystyle{ieeenat_fullname}
    \bibliography{main}
}

\clearpage
\setcounter{page}{1}
\maketitlesupplementary

Sec~\ref{sec:details} provides implementation details. Sec~\ref{sec:liedown} demonstrates the extensibility of our system by incorporating a new interaction skill of lying down. 

\section{Implementation Details}
\label{sec:details}

\textbf{Physics Simulation} We adopt Isaac Gym~\cite{makoviychuk2021isaac}, a large-scale GPU-based parallel physical simulator developed by NVIDIA. The simulation runs at $60$ Hz while the control policy runs at $30$ Hz.

\vspace{2mm}
\noindent \textbf{Policy Training} The policy network, value network, and motion discriminator are constructed as three separate MLPs, and the hidden dimension of each MLP is $(1024, 512)$. ReLU activations are used. The value of the diagonal covariance matrix is set to 0.055. Proximal policy optimization (PPO) is adopted as the training algorithm. Training parameters are presented in Table~\ref{tab:hyperpara}. Training each control policy using PPO and IsaacGym takes $\sim$ 12 hours on a single NVIDIA Tesla V100 GPU.

\begin{table}[h]
  \centering
  \resizebox{0.6\linewidth}{!}{
    \begin{tabular}{c | c}
      \noalign{\hrule height 1pt}
      Parameter & Value \\
      \hline
      number of environments & 6144 \\
      batch size for PPO & 16384 \\
      batch size for AMP & 4096 \\
      horizon length & 32 \\
      learning rate & 5e-5 \\
      clip range $\epsilon$ for PPO & 0.2 \\
      discount factor $\gamma$ & 0.99 \\
      GAE coefficient $\lambda$ & 0.95 \\
      \noalign{\hrule height 1pt}
      
    \end{tabular}
  }
  
  \caption{Training parameters of PPO and IsaacGym.}
  \vspace{-6mm}
  \label{tab:hyperpara}
\end{table}

\section{Learning to Lie Down}
\label{sec:liedown}

We train a new interaction controller consisting of two control policies capable of lying down and getting up, respectively. The controller can be seamlessly integrated into our system as a new low-level executor, providing more optional actions for the finite state machine planner. The training process of the two control policies is similar to Sec~\ref{sec:method_intercon}. We mainly introduce the differences here.

\vspace{2mm}
\noindent \textbf{Motion and Object Datasets} We use $31$ motion sequences from the SAMP dataset~\cite{hassan2021stochastic}. We select $57$ sofas from the 3D-Front object dataset~\cite{fu20213d} and randomly divide $50$ as the training set and $7$ as the testing set.

\begin{figure}[h]
  \centering
   \includegraphics[width=0.7\linewidth]{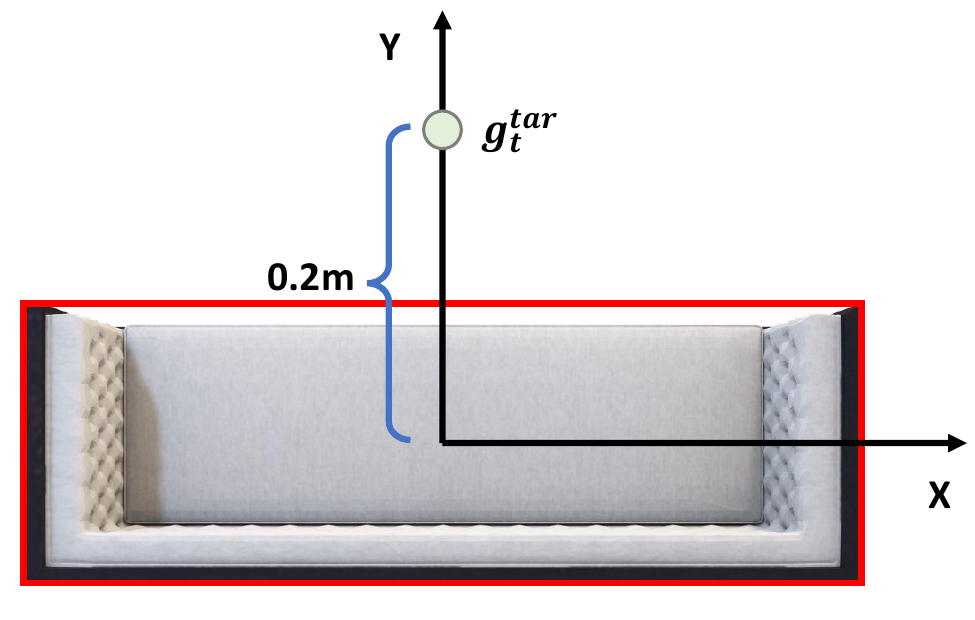}
   \vspace{-4mm}
   \caption{Illustration about how to set the XY components of $g_t^{tar}$ in Equ~\ref{eq:far_reward_liewdown_getup} which is used during the training of the get-up policy.}
   \label{fig:getup_tar_pos}
\end{figure}

\vspace{2mm}
\noindent \textbf{Task Rewards} To train the lie-down policy, the construction of the total task reward is the same as Eq.~\ref{eq:sit_total_reward}. We only modify the equation of $r_t^{near}$ to better learn the final lying states. The new $r_t^{near}$ is defined as:
\begin{equation}
\text{exp}\big(-10.0 ( \left \| g_t^{tar} - x_t^{root} \right \|^2  + \left \| g_t^{tarHeadH} - x_t^{head} \right \|^2 ) \big)
\label{eq:near_reward_liedown}
\end{equation}
where $x^{head}_t$ is the height of the character's head, and $g_t^{tarH}$ is the target height. Specifically, the value of $g_t^{tarHeadH}$ is equal to the height component of $g_t^{tar}$. To train the get-up policy, we exploit a new total task reward $r^G_t$, as follows:
\begin{equation}
    \begin{aligned}
    r^G_t &= 0.5 \ \text{exp}\big(-10.0 \left \| g_t^{tar} - x_t^{root} \right \|^2 \big) \\ 
    &+ 0.3 \ \text{exp}\big(-10.0 \left \| g_t^{tarFootH} - x_t^{foot} \right \|^2 \big) \\ 
    &+ 0.2 \ \text{exp}\big(-10.0 \left \| g_t^{tarHeadH} - x_t^{head} \right \|^2 \big).
    \end{aligned}
\label{eq:far_reward_liewdown_getup}
\end{equation}
The height component of $g_t^{tar}$ is always equal to $0.89$, and the rest components that decide the 2D location of $g_t^{tar}$ are illustrated in Fig~\ref{fig:getup_tar_pos}.

\begin{figure}[h]
  \centering
   \includegraphics[width=0.9\linewidth]{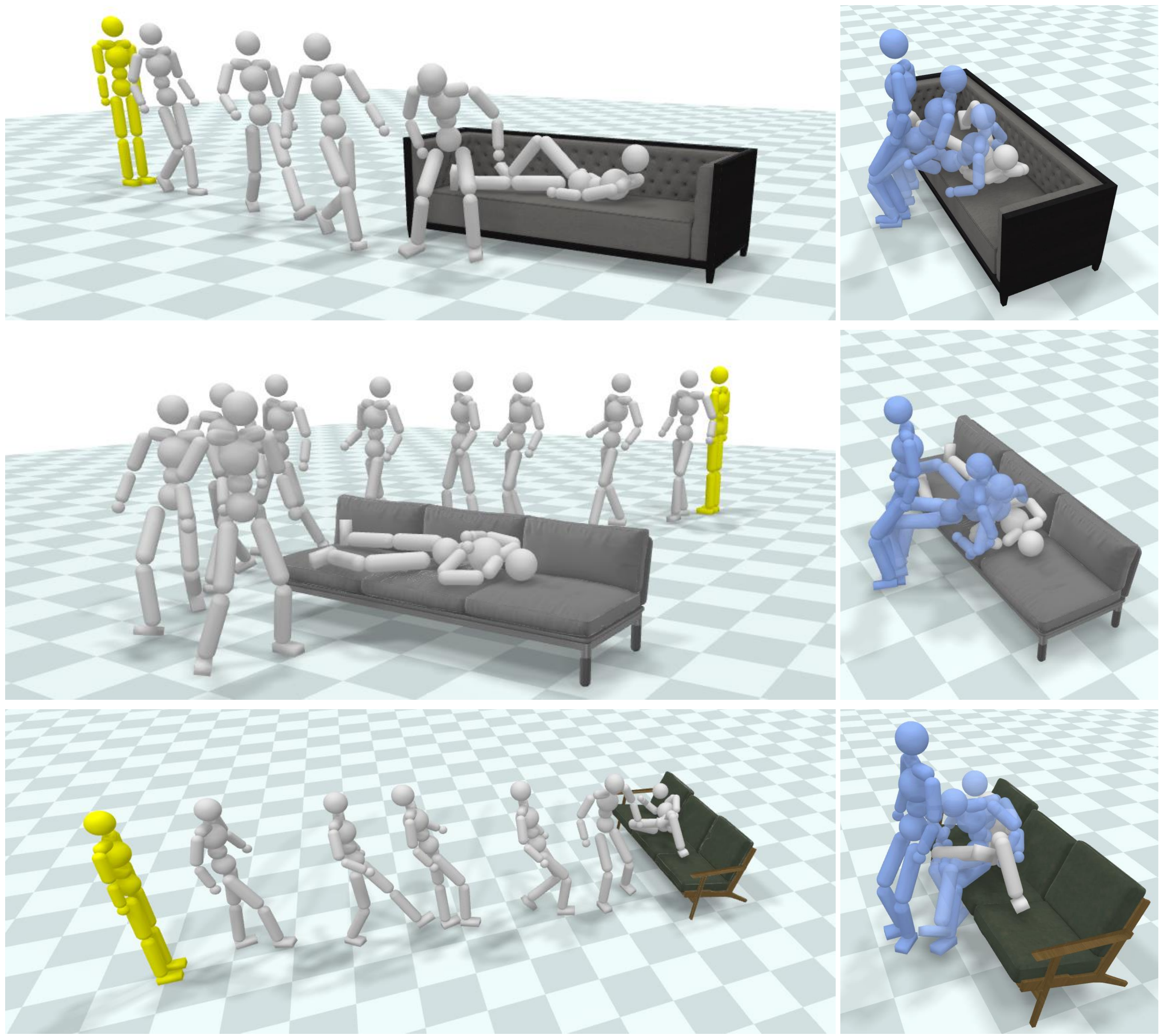}
   \vspace{-2mm}
   \caption{Character animations generated by the lie-down interaction controller..}
   \label{fig:liedown}
\end{figure}

\begin{figure*}
  \centering
   \includegraphics[width=1.0\linewidth]{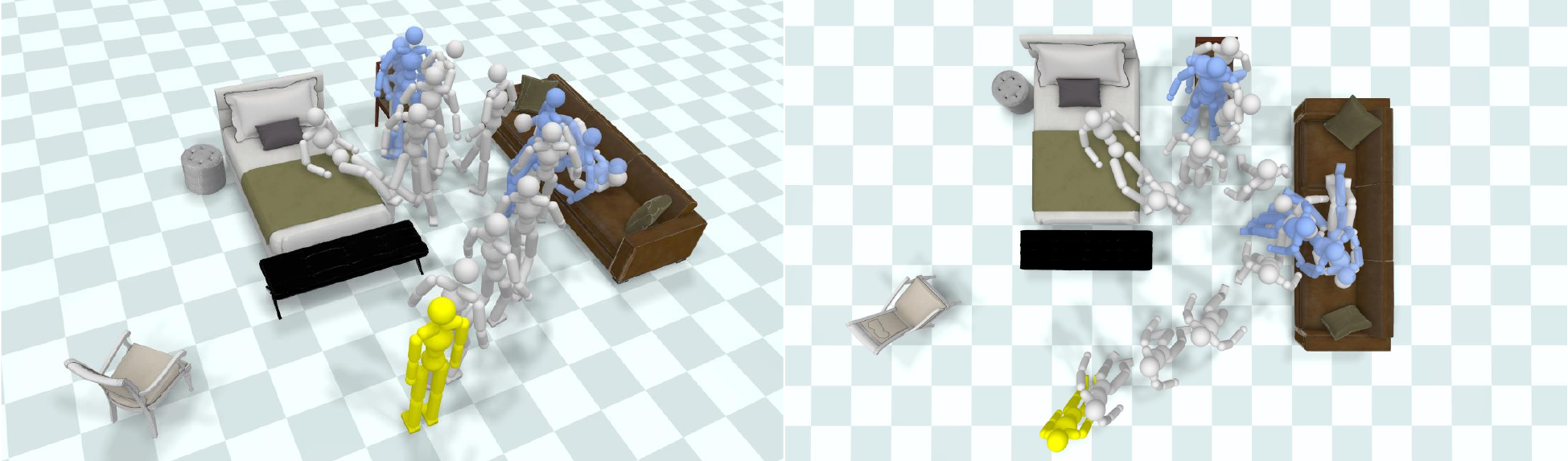}
   \caption{Given two interaction controllers, our extended system enables the physics-based character to first lie on the sofa, then sits on the chair, and finally lie on the bed, exhibiting more diverse long-term interactions.}
   \label{fig:scene_liedown}
\end{figure*}

\vspace{2mm}
\noindent \textbf{Results} As illustrated in Fig~\ref{fig:liedown}, using the new interaction controller, we can control the physics-based character to lie down and get up from large sofas naturally and successfully. Furthermore, we can seamlessly integrate the new interaction controller into the existing system. In such a case, the finite state machine planner is able to schedule three low-level executors. As shown in Fig~\ref{fig:scene_liedown}, the extended system can synthesize more diverse long-term motion sequences in 3D indoor scenes, which demonstrates the strong extensibility of our framework.

\end{document}